%% file: icb19.tex
\documentclass[10pt,twocolumn,letterpaper]{article}

\usepackage{icb}
\usepackage{times}
\usepackage{epsfig}
\usepackage{graphicx}
\usepackage{amsmath}
\usepackage{amssymb}
\usepackage{multirow}
\usepackage{iac_pkg}
\usepackage{tabularx, booktabs}
\usepackage{subcaption}
\usepackage{wrapfig}
\usepackage{footmisc}
\usepackage[labelfont={small},font=small]{caption}

\usepackage[pagebackref=true,breaklinks=true,colorlinks,bookmarks=false]{hyperref}

\icbfinalcopy %

\makeatletter 
\def\ps@IEEEtitlepagestyle{ 
\def\@oddfoot{\mycopyrightnotice} 
\def\@evenfoot{} 
} 
\def\mycopyrightnotice{ 
{\hfill \footnotesize 978-1-7281-3640-0/19/\$31.00 \copyright 2019 IEEE} 
} 
\makeatother 

\begin{document}

\title{Does Generative Face Completion Help Face Recognition?}
\author{Joe Mathai\textsuperscript{*}, Iacopo Masi\textsuperscript{*}, Wael AbdAlmageed\\
USC Information Sciences Institute, Marina del Rey, CA, USA\\
{\tt\small \{jmathai,iacopo,wamageed\}@isi.edu}
}

\maketitle
\thispagestyle{empty}

\input{sections/00_abstract.tex}
{\let\thefootnote\relax\footnotetext{{\footnotesize\textsuperscript{*} indicates equal contribution}}}
\input{sections/01_introduction.tex}

\input{sections/02_review.tex}

\input{sections/03_completion.tex}

\input{sections/04_testing.tex}

\vspace{-5pt}
\input{sections/05_conclusion.tex}
\vspace{0.5pt}

\input{sections/06_appendix.tex}

\section*{Acknowledgements:} This research is based upon work supported by the Office of the Director of National Intelligence (ODNI), Intelligence Advanced Research Projects Activity (IARPA), via IARPA R\&D Contract No. 2017-17020200005. The views and conclusions contained herein are those of the authors and should not be interpreted as necessarily representing the official policies or endorsements, either expressed or implied, of the ODNI, IARPA, or the U.S. Government. The U.S. Government is authorized to reproduce and distribute reprints for Governmental purposes notwithstanding any copyright annotation thereon.

{\small
\bibliographystyle{ieee}

}

\end{document}

%% file: sections/00_abstract.tex
\begin{abstract}
    Face occlusions, covering either the majority or discriminative parts of the face, can break facial perception and produce a drastic loss of information. Biometric systems such as recent deep face recognition models are not immune to obstructions or other objects covering parts of the face. While most of the current face recognition methods are not optimized to handle occlusions, there have been a few attempts to improve robustness directly in the training stage. Unlike those, we propose to study the effect of generative face completion on the recognition. We offer a face completion encoder-decoder, based on a convolutional operator with a gating mechanism, trained with an ample set of face occlusions.
    To systematically evaluate the impact of realistic occlusions on recognition, we propose to play the occlusion game: we render 3D objects onto different face parts, providing precious knowledge of what the impact is of effectively removing those occlusions. Extensive experiments on the Labeled Faces in the Wild (LFW), and its more difficult variant LFW-BLUFR, testify that face completion is able to partially restore face perception in machine vision systems for improved recognition.
\end{abstract}

%% file: sections/01_introduction.tex
\section{Introduction}\label{sec:intro}
Facial occlusions, whether for the majority of the face or only a few discriminative regions, alter the perception of the face and cause a drastic loss of information. Biometric systems, such as presentation attack detection (i.e. anti-spoofing) methods~\cite{Liu_2018_CVPR, Jourabloo_2018_ECCV} and face recognition systems ~\cite{taigman2014deepface,cao2017vggface2} are not immune to those occlusions. Common objects, such as sunglasses, hands, microphones, scarves, and hats could be purposely placed onto the face by a malicious attacker, producing occlusions with the intent of impairing the biometric system, possibly leading to significant performance degradation, unwanted authentications or spoofing attacks.

The biometrics and vision communities have made significant efforts to develop robust occlusion-resistant face analysis algorithms such as landmark detectors resilient to partial occlusions~\cite{burgos2013robust,ghiasi2014occlusion,yu2014consensus}, or face detection networks with occlusion-aware loss functions~\cite{opitz2016grid}. However, less emphasis has been given to studying the impact of occlusions on recognition systems.
Similarly, face occlusions, and their ultimate effect on face recognition, are often unexplored in the biometric community. 
This issue calls for biometric methods that are in principle robust to occlusions, or at least aware of malicious occlusions,  when they happen. 
In light of this, there have been a few attempts to directly improve face recognition robustness --- either by designing hand-crafted features that inhibit the occlusions in the feature space~\cite{jia2008face,lopezinhibition} or by using attention mechanisms to guide the training of convolutional neural networks to avoid occlusions ~\cite{wan2017occlusion}. Other attempts employed stacked denoising autoencoders for restoring the masked face parts~\cite{zhang2013occlusion,cheng2015robust}.

\begin{figure}[t]
\centering
\includegraphics[width=\linewidth]{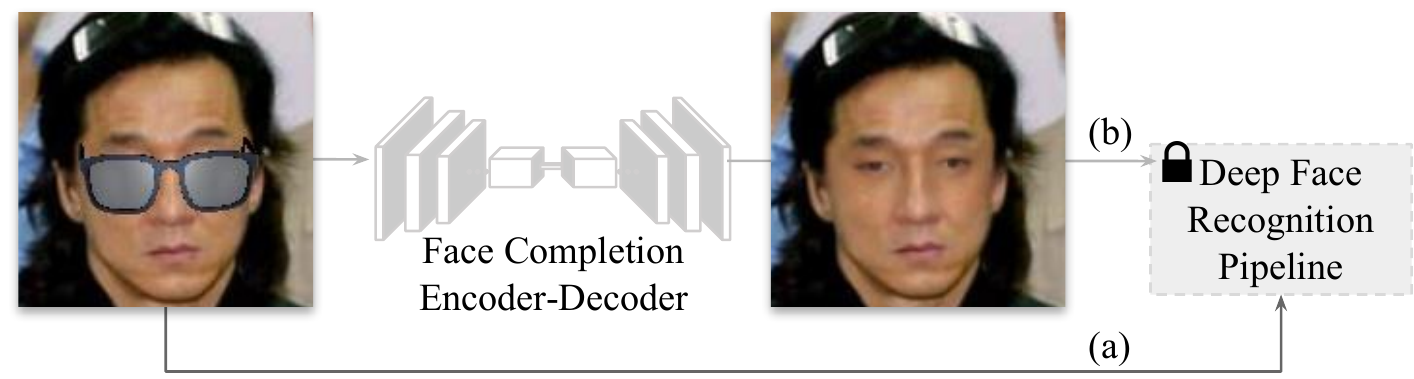}
\caption{
A realistically occluded face and its completed version by our face inpainter. We show the effect of face inpainting on the performance of a deep face recognition pipeline with occlusions. Note that we assume \emph{no} access to the recognition pipe. Our study aims to quantitatively assess (a) the impact of different synthetic yet realistic occlusions on recognition and  (b) how much face perception is restored via face inpainting.
\vspace{-15pt}}
\label{fig:teaser}
\end{figure}

Unlike previously mentioned methods, we take an orthogonal approach to handle face occlusions: instead of optimizing a face recognition network for occlusion robustness, we leverage deep face completion networks ~\cite{li2017generative,iizuka2017globally}, and conduct a novel study on the effect of generative face completion --- also known as face inpainting --- on the performance of state-of-the-art face recognition systems, as illustrated in  \figref{teaser}. In doing so, we attempt  answering the following questions: \emph{(1) how does a generative face completion result interact with the performance of a deep face recognition pipeline? (2) Referring to \figref{teaser}, does the scenario (b) provide substantial improvement over (a)? (3) What is the quantitative performance difference of diverse occluding objects and by how much is the inpainter able to cope with them?}
Furthermore, unlike previous work, we make an effort to factor out the impact of occlusions produced by a diverse set of realistic objects, without just using rectangles unrealistically synthesized on the face. We argue that this latter form of occlusion (i.e. rectangular occlusions) is rather too simple to find in real scenarios, and hence, studies conducted on this type of occlusions might be too unrealistic for drawing conclusions.
Finally, we  provide a systematic and rigorous evaluation pipeline tested on public benchmarks of face recognition in the wild. Since it has recently been demonstrated that Labeled Faces in the Wild (LFW) verification performance has been saturated~\cite{taigman2014deepface}, we further extend our study to the more difficult variant, LFW Benchmark of Large-scale Unconstrained Face Recognition (BLUFR)~\cite{liao2014benchmark}. Experimental evaluation verifies that face completion partially restores face perception of machine vision systems for improved recognition under mild occlusions, though exhibits less benefits for strong occlusions on the image boundary.
Technically we focus on an effective face completion network and novel evaluation framework to study the impact of occlusions on faces in-the-wild. %

This paper has three main contributions. First, we offer an effective face completion encoder-decoder that inpaints real object occlusions (not just rectangles) and deals with holes on discriminative parts of the face. The encoder-decoder is trained with an abundant set of diverse masks based on both regular and irregular occlusions, tailored to the face domain. To propagate the semantics of the generated masks to the entire network all the way down to the final loss, we study the effects of convolutions along with a gating mechanism~\cite{liu2018image,dauphin2016language,yu2018free}. Second, we propose to ``play the occlusion game,'' which means to realistically cover discriminative parts of the face with synthetic yet realistic objects through a rendering mechanism and register recognition performance before and after face completion. Third, despite recent work on deep face inpainting~\cite{li2017generative,chen2018high,zhaoidentity,yeh2017semantic}, to the best of our knowledge we are the first to quantitatively test its role in a biometric system. 
To enable reproducibility of our results,we publicly release our code and models\footnote{Available at \href{https://github.com/isi-vista/face-completion}{github.com/isi-vista/face-completion}}.

The paper is organized as follows. In \secref{related} we review recent work on face completion, general-purpose inpainting, and finally occlusion-robust face recognition. \secref{inpainter} introduces our encoder-decoder for face completion in-the-wild. We explain the evaluation framework along with gained insights from the results in \secref{expt}. We draw conclusions and discuss future work in \secref{conclusions}.

%% file: sections/02_review.tex
\section{Related Work}\label{sec:related}

\minisection{Face Completion ---} Generally speaking, to solve the face completion task, a generative network is trained with paired occluded and unoccluded faces. Previous face completion methods include an encoder-decoder architecture to hallucinate the occluded face part~\cite{li2017generative}. The system is trained with na\"ive synthetic occlusions (square masks), and uses global and local adversarial losses~\cite{iizuka2017globally} along with a face-parsing loss to enforce face symmetry in the reconstruction. While the previous method works with tiny input faces (64 or 128 pixels), a  recent method applied progressive growing GAN for high-resolution face completion~\cite{chen2018high} showing stunning high-resolution results. Zhao et al. \cite{zhaoidentity} fills occluded parts under large occlusions and pose variations, often occurring when wearing VR/AR devices. The system uses an identity loss for preserving characteristic traits of the individual in the reconstructed face. Song \etal~\cite{song2018geometry} designed a network to estimate the coarse geometry of the face, which is then used to condition the generator, which in turn disentangles the mask from the completed face image.

\minisection{Inpainting ---} Iizuka \etal~\cite{iizuka2017globally} proposed both local and global adversarial loss function for effective scene inpainting. These two losses ensure local and global realism of the reconstructed image. Liu \etal ~\cite{liu2018image} uses partial convolutions for the inpainting problem to handle irregular masks. Though partial convolutions were first introduced in~\cite{liu2015sparse}, and also used as attention masks in~\cite{harley2017segmentation}, they propose a novel update scheme for the mask that is partially reduced and propagated in subsequent deeper layers.
While learning-based methods need information about the mask at training time, it is noteworthy to mention the work in~\cite{yeh2017semantic}, which first trains a GAN to learn to generate from the real, unoccluded face appearance distribution. Then, given the learned manifold embedding, the method searches for the closest encoding given the corrupted input, using  context and prior losses. The retrieved encoding is then passed through the decoder to recover the occluded content.

\minisection{Face Recognition ---} Face analysis is an essential branch of computer vision, with different tasks falling under its broad scope, with work dating back to the sixties~\cite{bledsoe1966man}. Recently, face recognition made a  performance leap through the application of deep convolutional neural networks trained on huge, labeled training sets~\cite{parkhi2015deep} or with unique models that bundle all the face-related tasks into a single solution using multi-task learning~\cite{ranjan2017all}. In recent years, while a plethora of advanced methods has been proposed to push the frontiers of recognition, much less attention has been given to particular face pairs that can fool machines~\cite{kushwaha2017disguised_faces} or to understanding which type of occlusions negatively affect a system.
Despite over half a decade of deep face recognition~\cite{sun2014deep,taigman2014deepface}, there are still efforts  to unravel \emph{how and why} these models perform predictions, thereby understanding which visual facial cues stimulate certain activations in a neural network~\cite{castanon2018visualizing}. 
Previous work attempted to directly improve face recognition robustness to occlusions by designing  hand-crafted features (such as local binary patterns) that inhibit the non-face part in the feature space~\cite{jia2008face,lopezinhibition}. The same concept is later used for designing a convolutional network that learns attention masks to avoid computing convolution responses under the occluded parts~\cite{wan2017occlusion,trigueros2018enhancing}. Other approaches have employed stacked denoising autoencoders for restoring the masked face parts~\cite{Iliadis2017RobustAL,zhang2013occlusion,cheng2015robust}.%

%% file: sections/03_completion.tex
\section{Encoder-Decoder for Face Completion }\label{sec:inpainter}
A key component for developing an effective face inpainter relies on having a clean (i.e. unoccluded) face dataset that can be used as a basis set for generating synthetic face occlusions for learning the face completion task. However, harvesting faces in the wild does not satisfy this requirement since those could be contaminated by occlusions and other obstructions. In order to properly supervise the proposed face completion encoder-decoder, we leverage large face collections borrowed from the face recognition community, where data is abundant. We used images from CASIA WebFaces~\cite{yi2014learning}, VGG Faces~\cite{parkhi2015deep} and MS-Celeb-1M~\cite{guo2016ms} datasets for training the encoder-decoder inpainter. To mitigate the presence of occlusions, we initially run a face detector~\cite{yang2016multi} and retain a face image for training if and only if the face detector is highly confident of its prediction (confidence $\geq 0.985$). This process leads to a training data of 197,081 clean images, while maintaining most of the tough variability present in the original data sets. Another 21,121 validation images are used for hyperparameter tuning. Faces are aligned by expanding the detector box into a square from its center to compensate for scale and translation.

\subsection{Occlusion Mask Synthesis}\label{sec:occ_mask}
\begin{figure}[tb]
    \centering
    \begin{subfigure}[t]{0.25\linewidth}
        \includegraphics[width=\linewidth]{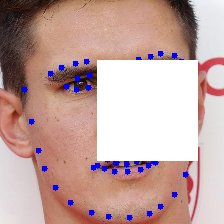}
        \caption{}\label{fig:masks:a}
    \end{subfigure}\qquad
   \begin{subfigure}[t]{0.25\linewidth}
        \includegraphics[width=\linewidth]{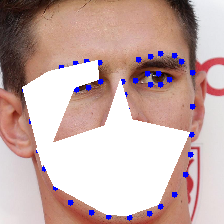}
        \caption{}\label{fig:masks:b}
    \end{subfigure}\qquad
   \begin{subfigure}[t]{0.25\linewidth}
        \includegraphics[width=\linewidth]{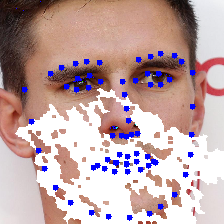}
        \caption{}\label{fig:masks:c}
    \end{subfigure}
   \begin{subfigure}[t]{0.25\linewidth}
        \includegraphics[width=\linewidth]{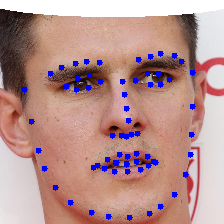}
        \caption{}\label{fig:masks:d}
    \end{subfigure}\qquad
   \begin{subfigure}[t]{0.25\linewidth}
        \includegraphics[width=\linewidth]{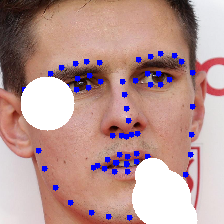}
        \caption{}\label{fig:masks:e}
    \end{subfigure}\qquad
   \begin{subfigure}[t]{0.25\linewidth}
        \includegraphics[width=\linewidth]{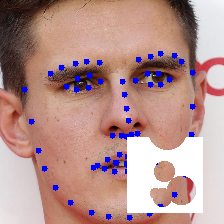}
        \caption{}\label{fig:masks:f}
    \end{subfigure}
    \caption{Occlusion mask synthesis used for training along with landmarks: (a) a square crop centered on a random fiducial point; (b) irregular masks based on a pool of combinations of landmarks defined a priori; (c) random-walk masks, similar to~\cite{liu2018image} (d) boundary masks (e) holes (f) eroded crops.}
  \label{fig:masks}
  \vspace{-10pt}
\end{figure}

The encoder-decoder is trained with different masks covering the face, of both regular and irregular shapes. We tailored the training occlusion masks to the face domain by proposing different occlusions: given a 128$\times$128 input face image, it is obstructed by regular square crops of 50 pixels expanded from a fiducial point randomly sampled from detected landmarks on the face, as shown in \figref{masks} (\subref{fig:masks:a});  irregular landmark-based masks defined by connecting different landmarks.  These masks are sampled from a pool of predefined combinations of landmarks, often occluding sensible parts of the face --- eyes, nose, mouth as reported in  \figref{masks} (\subref{fig:masks:b}); random-walk masks usually used in face editing applications~\cite{liu2018image} --- less relevant here but still useful for providing irregular patterns, see  \figref{masks} (\subref{fig:masks:c}); masks on the boundaries as depicted in  \figref{masks} (\subref{fig:masks:d}); holes and eroded crops depicted respectively in \figref{masks} (\subref{fig:masks:e}) and  \figref{masks} (\subref{fig:masks:f}).
Landmarks are generated following the head pose estimation method~\cite{chang17fpn}, projecting 3D landmarks of a generic model onto the face image. Though these landmarks are not extremely accurate, they are much more robust than landmark detectors, and their accuracy is enough to effectively generate the occlusions in the wild, shown in \figref{masks}.

\subsection{Inpainting Network and Gated Convolution} \label{sec:net_inp}
We used an encoder-decoder based on the U-Net architecture~\cite{isola2017image}, but without using skip connections, and using convolutions with a gating mechanism along with substantial modification described below. In particular, we experiment with two different ways to implement gated convolutions --- hard gating mechanism, also known as partial convolution~\cite{liu2018image}, and the soft version based on a sigmoid function ~\cite{dauphin2016language,yu2018free}.
Our architecture consists of a generator $\phi_G$ and two discriminators: a patch-based local discriminator $\phi_{D_{l}}$ and a global, holistic $\phi_{D_{g}}$. The discriminators are only used at training time to enforce realism of the generated data, and are discarded at test time.

\minisection{Encoder-Decoder Architecture:} The encoder consists of four blocks of convolutions with a gating mechanism, increasing the feature depth up to 256. After that, we have another four encoding blocks but with dilation equal to two. We do not use any pooling layers, and the spatial dimensionality of feature maps is reduced by striding. The decoder has the inverse structure of the encoder but uses nearest neighbour interpolation layers to up-sample the feature maps. We use upsampling instead of deconvolution to prevent checkerboard artifacts. A regular convolutional layer and hyperbolic tangent operator are used to synthesize the completed face image.

\minisection{Training:}
We train all our models with $\ell_1$ reconstruction loss along with two adversarial losses using two discriminators $\phi_{D_{l}}$ and $\phi_{D_{g}}$. %
The global adversarial loss used is the improved Wasserstein loss~\cite{gulrajani2017improved} with gradient penalty, as shown in \eqnref{training}:
\begin{multline}
    \mathop{\mathbb{E}}_{\tilde{{y}} \sim \mathbb{P}_g} \lbrack \phi_D \big(\tilde{\mbf{y}}\big) \rbrack -
    \mathop{\mathbb{E}}_{y\sim \mathbb{P}_r} \lbrack \phi_D \big( \mbf{y} \big) \rbrack + \\ +
    \lambda\mathop{\mathbb{E}}_{\hat{{y}} \sim \mathbb{P}_{\hat{{y}}}} \lbrack \big( \parallel \Delta_{\hat{\mbf{y}}} \phi_D\big(\hat{\mbf{y}}\big) \parallel_{2} - 1 \big)^2\rbrack
    \label{eq:training}
\end{multline}
where $\tilde{\mbf{y}} \doteq \phi_G(\mbf{x})$ is a generated sample, $\mbf{y}$ indicates a real image, and $\hat{\mbf{y}}$ is an interpolated sample between generated and real as mentioned in~\cite{gulrajani2017improved}. 
The patch-based discriminator follows~\cite{isola2017image} which restricts the attention to the structure in local image patches and classifies each patch as real or fake. Empirically, we have found that a patch discriminator with receptive field size of 50$\times$50 pixels in the input image produces realistic facial features. 
To stabilize the training of the discriminators along with the generator, we  increase the learning rate of the generator more than that of the discriminators. This is different than the regular policy that updates the generator more than the discriminator. The learning rate schedule follows roughly the cycle learning rate process mentioned in~\cite{smith2017cyclical}. 

\subsubsection{Gating Mechanism}
To shed some light on the the idea of conditioning the convolutional operator to avoid the occluded area, we perform ablations for different convolutional gating mechanisms. In particular, we implemented the state-of-the-art hard gating mechanism based on partial convolution~\cite{liu2018image}. Additionally, we compare it against its softer version that uses a sigmoid function to dynamically avoid the masked region. %

\minisection{Hard Gating:} Partial convolutions have been recently used in editing application for face completion in~\cite{liu2018image}. Here we test their effect in biometrics, where it is usually tougher for face completion since the inpainter needs to deal with low resolution images or images with harsh image quality. The idea is that the mask, fed as input to the network, is used to hardly gate the response of the convolutional operator, so that this latter does not take into account occluded pixels. Further details are given in~\cite{liu2018image}. Partial U-Net is the model using partial convolutions.

\minisection{Soft Gating:} Extending the previous concept, we changed the gating mechanism from a hard decision~\cite{liu2018image} to a soft gating function~\cite{dauphin2016language,yu2018free}. This acts as a softer version of the partial convolution to learn rich features optimized for face completion. The idea behind soft gating is that the input tensor $~\mbf{x}$ --- which can be in the first layer the input image alone or concatenated with the mask --- is processed by two sets of convolutional filters: regular filters  $\mbf{w}_f$ and gating filters $ \mbf{w}_g$; the latter are learned with a sigmoid function, acting as a soft-mask gating function. As described in \eqnref{gated_conv}, the output feature map $\mbf{x}^\prime$  is then obtained as
\begin{equation}
\mbf{x}^\prime = \mbf{w}_f~\mbf{x}^T \odot \sigma \big(\mbf{w}_g~\mbf{x}^T\big)~~\text{with}~~\sigma \in [0..1]
\label{eq:gated_conv}
\end{equation}
where $\odot$ represents the Hadamard product. Nonlinear activations based on Elu~\cite{clevert2015fast} are used after convolution $\mbf{w}_g$. The output feature map $\mbf{x}^\prime $ is normalized using instance normalization~\cite{vedaldi2016instance}. Gated U-Net uses gated convolutions of \eqnref{gated_conv} in all the layers\footnote{Gated convolutions by definition of \eqnref{gated_conv} double the number of parameters in a model since they require twice of convolutional filters.}.

%% file: sections/04_testing.tex
\section{Experimental Evaluation}\label{sec:expt}
In this section we show what  the effect is of meaningful occlusions with common objects that an intruder could use to fool a biometric system. We quantitatively asses the effect that generative face completion has on the recognition module to check at what extent it is possible to recover the lost information.
To this end, we used face imagery from Labeled Faces in the Wild (LFW~\cite{LFWTech}) and evaluate this on the standard 6,000 LFW verification pairs; we further test it on its much harder variant LFW-BLUFR~\cite{liao2014benchmark} for open-set identification, testifying that face completion is able to partially restore face perception in machine vision systems under mild occlusions.

\subsection{Occlusion Testing Pipeline}\label{sec:occ_testing}
\begin{wrapfigure}{r}{0.25\textwidth}
    \centering
    \hspace*{-5pt}%
    \includegraphics[width=\linewidth]{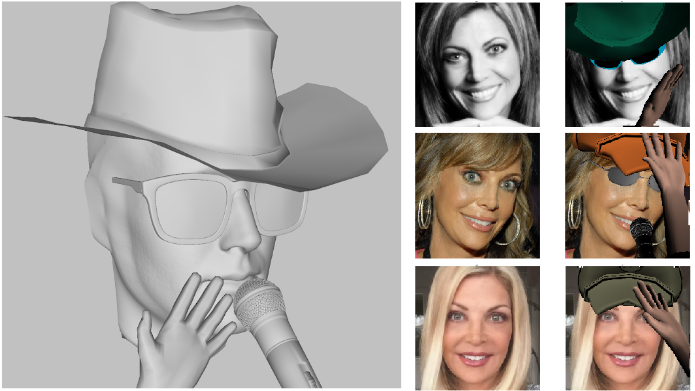}
    \caption{3D models and occlusion transfer onto face images.}
    \label{fig:3dsynth}
    \vspace{-5pt}
\end{wrapfigure}

In order to quantitatively measure the effect of face completion on recognition and control the side effects of face obstruction, we employed a synthetic yet realistic data rendering mechanism to add occlusions to an input face image. In contrast to  previous work~\cite{wan2017occlusion} that evaluates the effect of occlusion \emph{only} with square masks or that synthesize occlusions in 2D on frontal faces~\cite{zhao2018robust}, we offer a broader study with semantic occlusions of common obstructing objects properly places on the face (sunglasses, eyeglasses, microphones, etc.) through a robust 3D rendering mechanism described below. Note that all the introduced occlusions  are rendered on top of the occlusion naturally contained in the set. We defer the detection of those natural occlusions as an immediate future work (more on this in \secref{conclusions}).

\minisection{Realistic Occlusion Synthesis:} Inspired by domain randomization~\cite{Tremblay_2018_CVPR_Workshops}, we use a pool of 3D models along with their texture to render different objects onto a subject's face to measure their disruptive effect. \figref{3dsynth} shows a few examples without their texture along with the generic model, that is instead not rendered on the image. The same figure presents a few occlusions transferred onto the face images. In contrast to simply covering the face with square crops, our mechanism challenges the biometric recognition system by injecting occlusions similar to those of real scenarios. Our occlusions are rendered following the natural expected depth of objects---e.g. hands are farthest from the face, followed by caps, sunglasses and so on. %
In total we used twelve models of sunglasses, eyeglasses, microphone, hands, and cap or hats. This initial seed set is further expanded similarly to data augmentation but randomly sampling for plausible 3D variations in scale $s$, 3D rotation $R\in \mathbb{R}^3$, and translation $t\in \mathbb{R}^3$. The texture of the objects is also randomly changed. Finally, the 3D pose of the 3D generic model respect to the face image is estimated using~\cite{chang17fpn} and then used to render the object and register its occlusion mask.

\minisection{Face Recognition Engine:} We employ two state-of-the-art face recognition engines as a proxy to measure the impact of face completion to the final recognition accuracy. We use two engines ~\cite{masi2019facespecific,cao2017vggface2} to minimize the bias towards a particular recognition method. 
We emphasize that we used on purpose off-the-shelf recognition networks trained for the generic face recognition task, \emph{not explicitly optimized to deal with occlusions}. This better motivates our work and put all the burden on our face completion network, better isolating its impact.
The first recognition network is a ResNet-101 trained with face-specific data augmentation~\cite{masi2019facespecific}; we further experiment with the very recent VGGFace2~\cite{cao2017vggface2}.

\minisection{Face Recognition Pipeline:}
Faces are detected following~\cite{yang2016multi}. We also used FacePoseNet (FPN)~\cite{chang17fpn} to compute the 3D pose of each face yet aligning the faces with a simple 2D similarity transform to a reference face template. To deal with profile faces, we used a different coordinate system for frontal faces and profile faces. We should note that we apply the same preprocessing to both encoders, thus VGGFace2 is not optimized to this preprocessing step.
For both of the two encoders, faces are represented using the response of the layer prior the classification layer. %
The learned embedding is refined using principal component analysis (PCA) learned on each training split of the tested benchmarks. Power normalization is applied to the feature after projecting them with PCA.
Given an input face, the final representation is the result of average-pooling the 2D aligned face and its horizontal flip.
Face encodings are compared with cosine similarity. %

\subsection{Ablation Study}\label{sec:qual}
\begin{figure*}[tb]
    \centering
    \includegraphics[width=.95\linewidth]{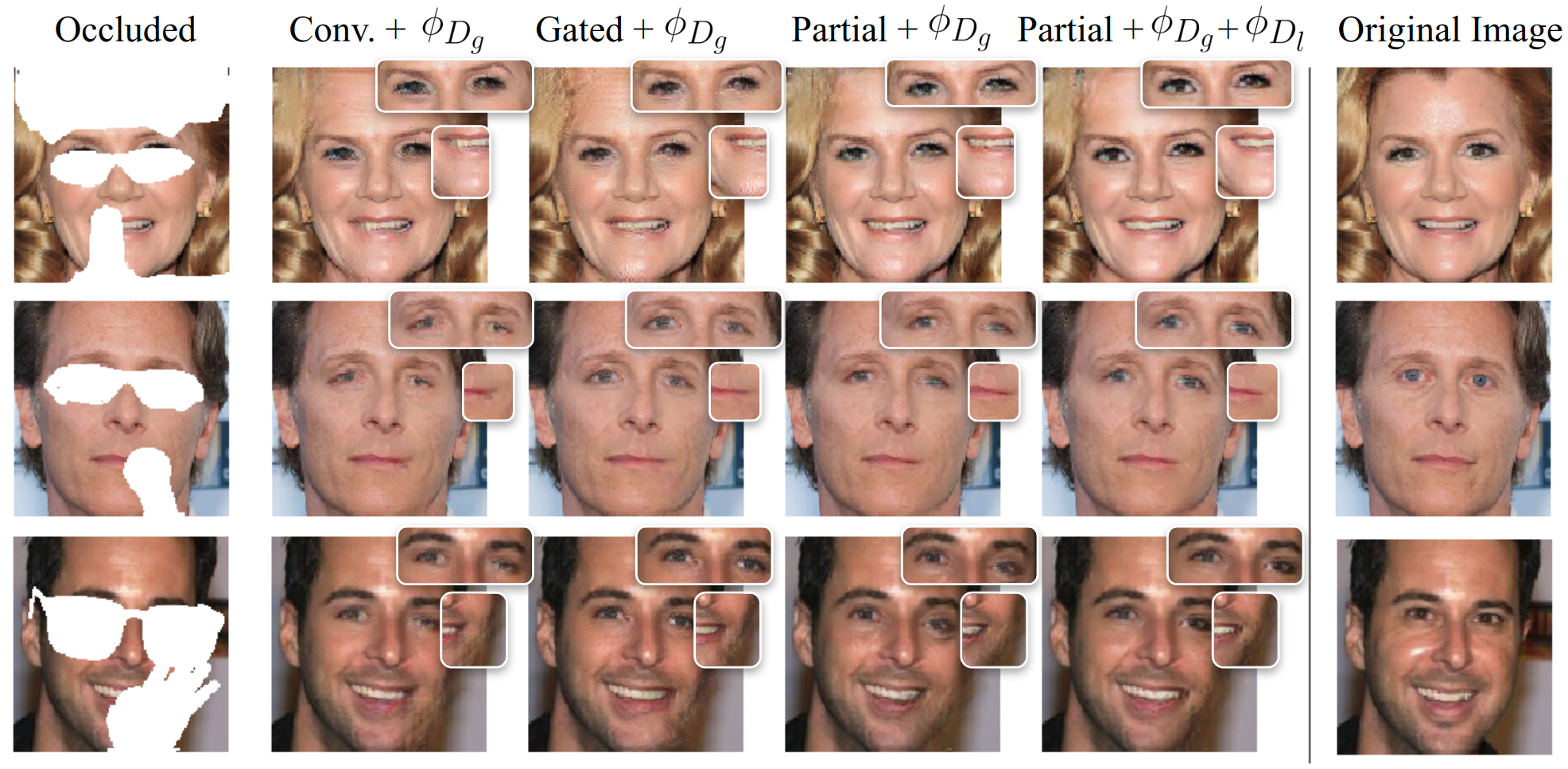}
    \caption{The effect of the gating mechanisms and the two discriminators: though the difference is subtle, the synthesis with gating contains less artifacts. Partial convolution in general performs better than gated convolution. The local discriminator further improves results.}
    \label{fig:qual_ab_b}
    \vspace{-5pt}
\end{figure*}

We conducted an ablation study using different gating mechanisms compared to classic convolutions. To the best of our knowledge, convolutions with soft gating are not popular for inpainting with the only exception of~\cite{yu2018free}. Though ~\cite{yu2018free}  reports improvement over classic convolution, it is instructive to study their generalization ability in the wild and compare it with the hard gating of~\cite{liu2018image}.

In order to marginalize out the effect of gating on the results, we also trained another model with the \emph{same} architecture of the Gated U-Net yet with classic convolutions. In addition, we used the same encoder-decoder but replaced soft gating with partial convolution.  We strove to deliver a fair comparison and trained the model for the same amount of time. Note that all the models tested are \emph{not} trained with the occlusions simulated at test time.

\begin{table}[tb]
\centering
\resizebox{\linewidth}{!}{
\setlength{\tabcolsep}{.21em}  %
\begin{tabular}{l c@{~}c@{~}c@{~}c@{~}c@{~}c@{~}}
\toprule
& & &  \multicolumn{4}{c}{\textbf{DIR\%@}}\\
\textbf{Method} & \textbf{Global}  & \tbf{Local} &FAR=100\% &FAR=10\% &FAR=1\% &FAR=0.1\%  \\
\cmidrule(r){1-1} \cmidrule(l){2-7}
Baseline & -- & -- & 97.28&	93.76&	82.69&	43.64\\
Sunglasses Occ. & -- & -- & 84.43&	67.99&	41.79&	17.58\\
\cmidrule(r){1-1} 
U-Net& \checkmark & -- & 88.85&	77.39&	54.37&	18.12\\
Gated U-Net& \checkmark & -- & 89.49&	77.21&	54.70&	19.04\\
Partial U-Net& \checkmark & -- & \tbf{89.70} &	 \tbf{78.50}&	 \tbf{57.78}&	23.00\\
\cmidrule(r){1-1}
Partial U-Net & \checkmark & \checkmark & 89.68&	78.33&	56.48&	 \tbf{25.16}\\
\bottomrule\end{tabular}
}
\caption{Ablation study on LFW BLUFR - identification: the use of different gating mechanism along with the effect of global and local discriminator. The results are reported on the ten splits as $\mu-\sigma$ using ResFace101~\cite{masi2019facespecific}.}
\label{tab:ablation}
\vspace{-5pt}
\end{table}

\minisection{Is gating necessary for training?} \figref{qual_ab_b} shows an ablation study from our validation set with \emph{multiple} masks \emph{unseen} in training. 
At a glance it can be noticed that \eqnref{gated_conv} helps the face completion task especially for difficult parts such as the eyes and the eyebrows as evident from \figref{qual_ab_b}, when compared with simple convolutions. Interestingly, though the Gated U-Net contains more parameters, Partial U-Net delivers the best result.
This supports our choice for involving a gating mechanism in the convolutional operator, with hard gating preferable over soft gating. The results with gating are in general overwhelmingly better than regular convolutions. We do observe a reduction on the amount of artifacts in the generated image that better restores the face manifold.
Additionally, in \figref{qual_ab_b}, all models use $\phi_{D_{g}}$. A marginal improvement is noticeable with the use of a local discriminator $\phi_{D_{l}}$.
Importantly, although here we show occlusion masks marked with white, the face recognition system is actually tested with realistic occlusions, as shown in \figref{3dsynth}.

We conclude that, in general, we can see that the gating mechanism helps the inpainting task better --- soft gating is better than simple convolution, though better results can be obtained with the hard gating implemented through partial convolution. \tabref{ablation} further supports  this claim. We conducted a face recognition experiment on LFW-BLUFR identification protocol injecting realistic sunglasses occlusions following \secref{occ_testing}. We performed this test with this type of occlusion since we know from recent studies~\cite{castanon2018visualizing}, aimed at understanding deep face recognition, that these models are very sensitive about the eye region. Our baseline recognition system obtains numbers near those of recently published systems~\cite{wang2018additive}.
\tabref{ablation} too proves the effectiveness of gating. The table shows the DIR (Detection and Identification Rate) at multiple False Positive Rates (FAR)~\cite{phillips2011evaluation}: though occlusions cause a drop in recognition, the encoder-decoder trained with the gating mechanism better restores the missing content, thereby improving face recognition accuracy. The recognition accuracy at rank-1, that in this case corresponds to DIR@FAR=100\%, is increased by 5\%. Significant improvement at very low FAR is provided by Partial U-Net. Performance is further increased with the local discriminator.

\subsection{Playing the Occlusion Game}
\begin{figure*}[tb]
\centering
  \begin{subfigure}[t]{.45\linewidth}
    \includegraphics[width=\linewidth]{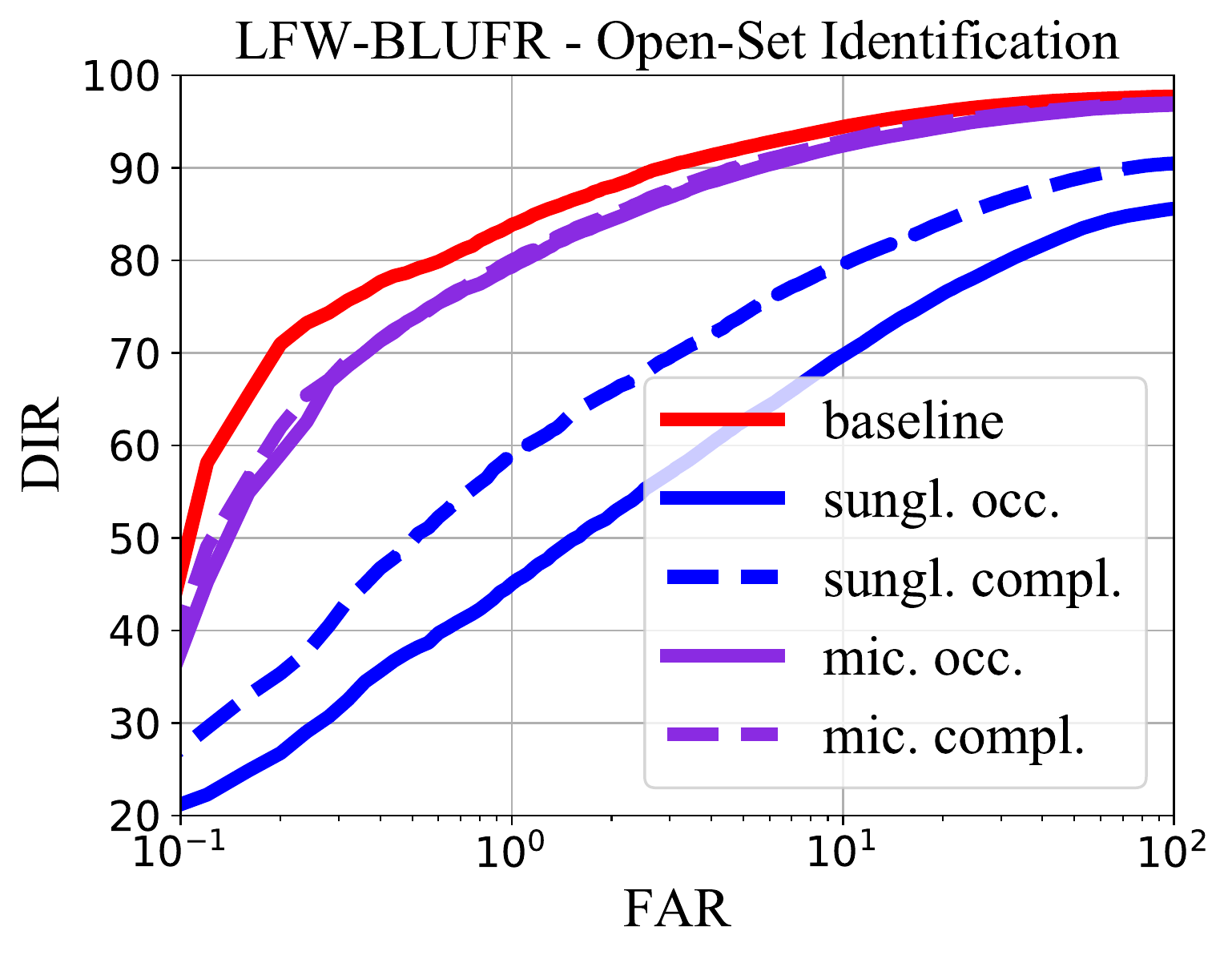}
    \caption{}\label{fig:dirs:a}
    \end{subfigure}\qquad
    \begin{subfigure}[t]{.45\linewidth}
    \includegraphics[width=\linewidth]{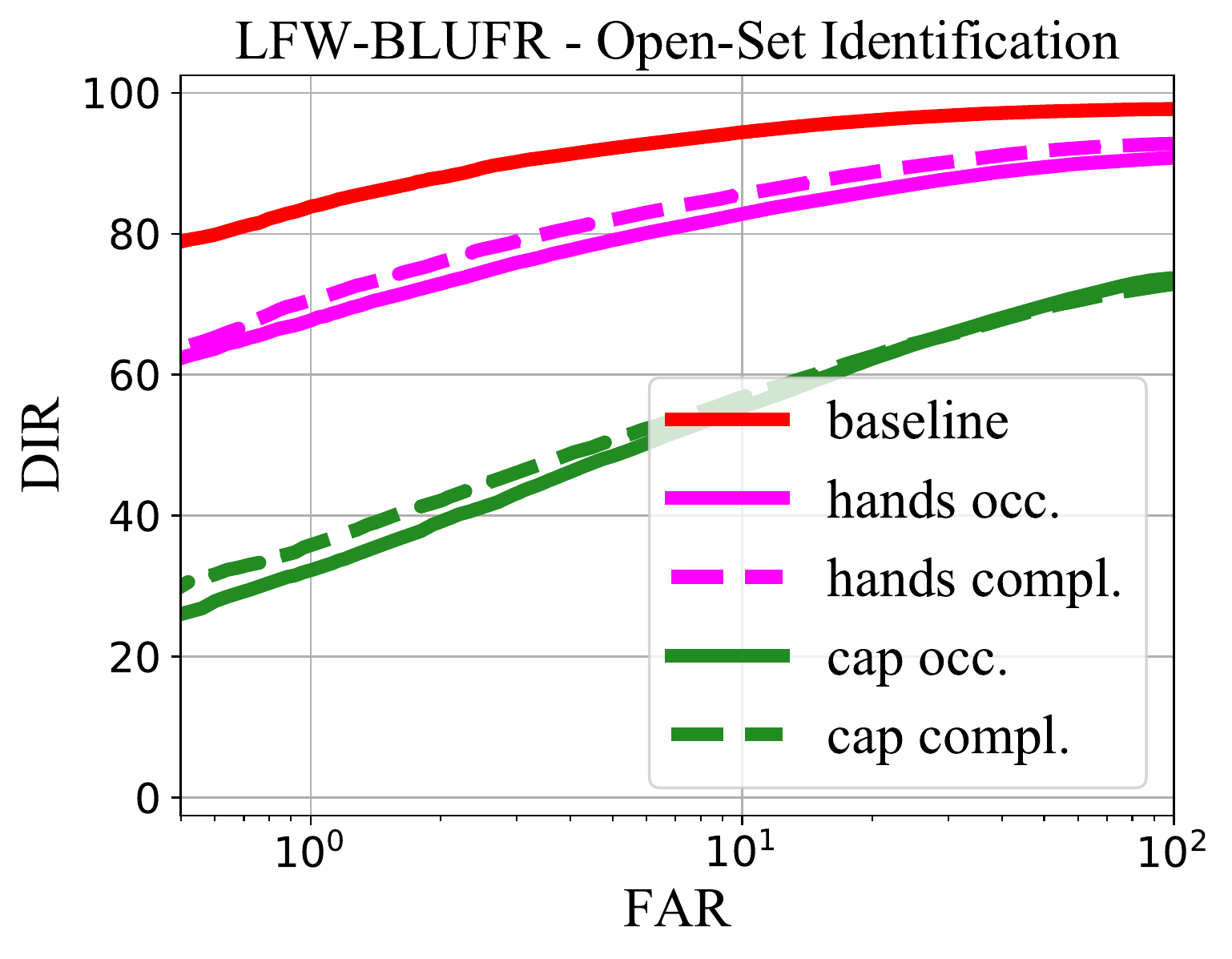}
    \caption{}\label{fig:dirs:b}
    \end{subfigure}
    \vspace{-5pt}
\caption{
LFW-BLUFR: recognition results~\cite{masi2019facespecific} and the effect of occlusions and generative face completion.
}
\label{fig:dirs}
\vspace{-10pt}
\end{figure*}

Following the takeaways of \secref{qual}, we quantitatively assess the effect of occlusions on a biometric face recognition system. We test if our best inpainter (Partial U-Net trained with two discriminators) can aid the recognition under occlusions. The bottom line idea is to simulate an intruder with the intent of fooling a recognition system so to avoid being recognized. This could be common in scenarios in which a system keeps a \emph{watch-list} of people to be monitored; the attacker may try to pass through the identification system occluding his face with multiple objects.
To perform a constructive study and control possible side effects, we propose ``to play the occlusion game'' which means realistically synthesizing plausible objects onto the face (\secref{occ_testing}) and observing the expected drop in recognition. This quantifies which parts of the face are more important for recognition and then evaluates if and how a generative face completion method helps to restore the lost information, thereby improving over possible obfuscations.
We report results of verification performance of the two different recognition engines (\secref{occ_testing}). Although we used different recognition systems, the takeaways from our study are consistent: in general, face completion helps restoring the lost information thereby improving recognition under mild occlusions; nevertheless the improvement is drastically reduced for boundary occlusions caused by hats or caps. The reason for this could be related to how the face completion system is trained with synthetic occlusion masks or simply because the majority of the face content is lost. \tabref{lfw:occlusions} reports the results for LFW verification, supporting our findings.
In order to challenge our experimental evaluation even more, we also offer the same study on the much harder LFW-BLUFR protocol for open-set identification, occluding only the probe set: this protocol fits best our setting since half of the probes is contained in the gallery (watch-list) while the other half is not: thus it provides a more general experimentation typically found in real scenarios.
\tabref{ablation} and \figref{dirs} show the DIR versus FAR~\cite{phillips2011evaluation} following the evaluation protocol for open-set recognition~\cite{liao2014benchmark}. \figref{dirs} supports the fact that generative face completion is beneficial for recognition for mild occlusions, such as those caused by sunglasses and eyeglasses; a small improvement at low FAR is reported for microphones, hands and hats.%

\begin{table}[tb]
\setlength{\tabcolsep}{2.5pt}
\centering
\resizebox{\linewidth}{!}{
\begin{tabular}{l ccccccc}
\toprule
Method & 1-EER  &Acc.&	AUC & & 1-EER  &	Acc. &	AUC  \\
\cmidrule(r){1-1} \cmidrule(l){2-4} \cmidrule(l){6-8}
& \multicolumn{3}{c}{ResFace101~\cite{masi2019facespecific}}& \multicolumn{3}{c}{VGGFace2~\cite{cao2017vggface2}} \\
Baseline& {99.43$\pm$.4} & {99.28$\pm$.4} & {99.96$\pm$.1} & & 98.37$\pm$.6& 98.25$\pm$.5& 99.71$\pm$.1  \\
\cmidrule(r){1-1}
\multicolumn{8}{c}{\tbf{Microphone}} \\
Occl. & 98.90$\pm$.5 & 98.85$\pm$.4 & 99.88$\pm$.1 &  & 97.63$\pm$.6& 97.68$\pm$.5& 99.53$\pm$.3\\
Compl. & 98.93$\pm$.6& 99.08$\pm$.5& 99.88$\pm$.1 &  & 97.77$\pm$.7& 97.87$\pm$.6& 99.68$\pm$.2\\
\cmidrule(r){1-1} \cmidrule(l){2-4} \cmidrule(l){6-8}
\multicolumn{8}{c}{\tbf{Hands}} \\
Occl. & 95.66$\pm$.9 & 96.05$\pm$.7 & 98.49$\pm$.5 &  & 92.53$\pm$1.& 93.10$\pm$1.& 96.94$\pm$.8\\
Compl. & 97.00$\pm$1.& 97.23$\pm$.7& 99.18$\pm$.5 &  & 95.13$\pm$1.& 95.53$\pm$.9& 98.56$\pm$.5\\
\cmidrule(r){1-1} \cmidrule(l){2-4} \cmidrule(l){6-8}
\multicolumn{8}{c}{\tbf{Sunglasses}} \\
Occl. & 96.90$\pm$.9 & 96.95$\pm$.9 & 99.19$\pm$.4 &  & 92.60$\pm$1. & 92.88$\pm$.9& 97.36$\pm$.4 \\
Compl. & 97.37$\pm$.7& 97.47$\pm$.6& 99.36$\pm$.4 &  & 94.77$\pm$1.& 95.12$\pm$.7& 98.63$\pm$.5\\
\cmidrule(r){1-1} \cmidrule(l){2-4} \cmidrule(l){6-8}
\multicolumn{8}{c}{\tbf{Cap}} \\
Occl. & 94.93$\pm$.4 & 95.25$\pm$.5 & 98.31$\pm$.3 &  & 91.33$\pm$1.& 91.82$\pm$.6& 96.44$\pm$.7\\ 
Compl. & 93.40$\pm$1.& 93.62$\pm$1.& 97.45$\pm$.7 &  & 91.70$\pm$2.& 92.37$\pm$.9& 96.96$\pm$.8\\
\bottomrule\end{tabular}
}
\caption{Playing the occlusion game on LFW verification. Results for different occlusions and the effect of face completion.%
}
\label{tab:lfw:occlusions}
\vspace{-7pt}
\end{table}

%% file: sections/05_conclusion.tex
\section{Conclusion and Future Work}\label{sec:conclusions}
In this paper we shed some light on the interactions between occlusions, inpainting and a biometric recognition system. We answered the questions that were raised in the title and \secref{intro}. Face completion seems beneficial for a biometric system, assuming that the occlusion can be detected effectively. On the other hand, for boundary occlusions, its contribution is minimal. %
As future work we will seek for a complete system to jointly \emph{detect and complete occlusions}. We will train the inpainter with occlusion masks transferred from a face segmentation method~\cite{nirkin2017face} and apply the latter to detect occlusions. Another interesting piece of future work is to employ a face verification loss to preserve the identity of the subject and to perform a thorough evaluation of completion algorithms.

%% file: sections/06_appendix.tex
\section{Appendix}
\subsection*{Inpainter Network Architecture Details (addendum to \secref{net_inp})}
\begin{figure*}[h]
\centering
\includegraphics[width=.85\linewidth, clip, trim=0mm 0mm 0mm 0mm]{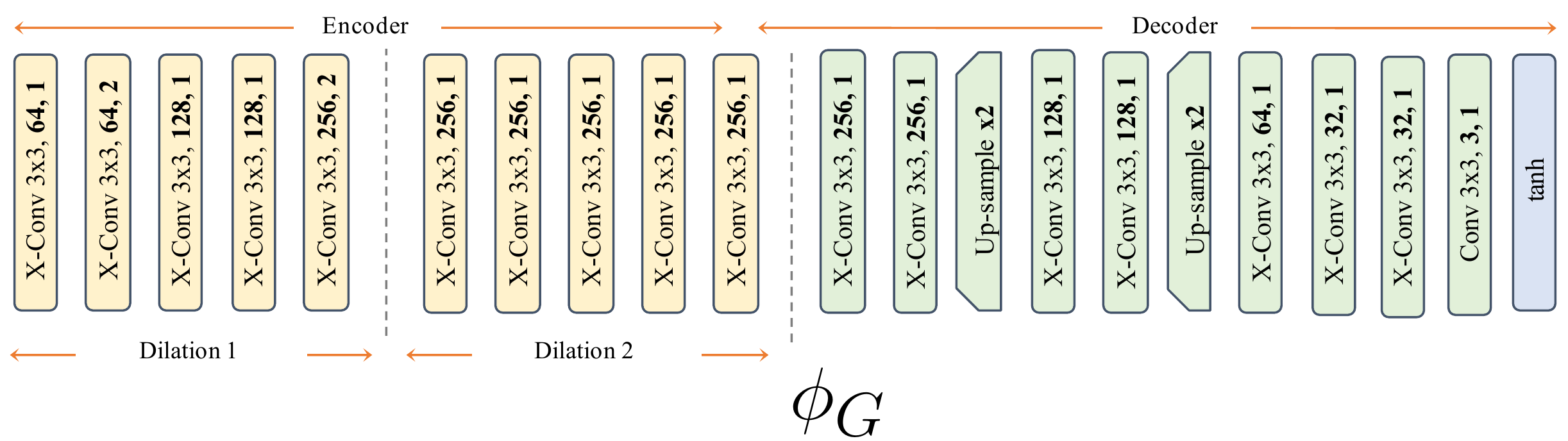}%
\caption{{\bf Encoder-Decoder Architecture Details.} We show the full architecture for the encoder-decoder. The encoding block is composed by two main parts: the first part uses convolution with dilation one; the second one uses dilation equal to two. The decoding block uses nearest neighbour interpolation for upsampling.}%
\label{fig:generator}
\end{figure*}

\minisection{Encoder-Decoder:} Our base encoder-decoder is inspired by U-Net from~\cite{isola2017image} but without skip connections. The input to the network is a 128$\times$128$\times$3 image concatenated with the mask along the channel dimension. Depending on the model type, we replace regular convolution (U-Net) with partial convolution (Partial U-Net) or gated convolution (Gated U-Net). The feature map size is decreased only using striding (i.e., {\em without} pooling layers). The architecture of the encoder-decoder is shown in \figref{generator}. Each convolutional layer is followed by Exponential Linear Unit (ELUs) activations. The output feature map is normalized using instance normalization~\cite{vedaldi2016instance}. 
The convolutional filters are initialized with the method~\cite{glorot2010understanding}; biases are initialized with a constant value of 0.0. Finally, the last decoding block has a regular convolutional layer and a hyperbolic tangent layer that map back to an output of dimension 128$\times$128$\times$3.
We used the same architecture for all the ablation study with the exception of replacing regular convolution with partial or gated convolution. In \figref{generator}, ``X-Conv'' indicates a placeholder for one among the three convolutions $\mathtt{\{regular, partial, gated\}}$. The legend for the blocks in \figref{generator} is as follows: $\mathtt{X-Conv 3x3,256,1}$ indicates convolutional filters with kernel 3x3, an output with 256 channels and stride equal to 1.

\begin{figure*}[h]
\centering
\includegraphics[width=.95\linewidth, clip, trim=0mm 0mm 0mm 0mm]{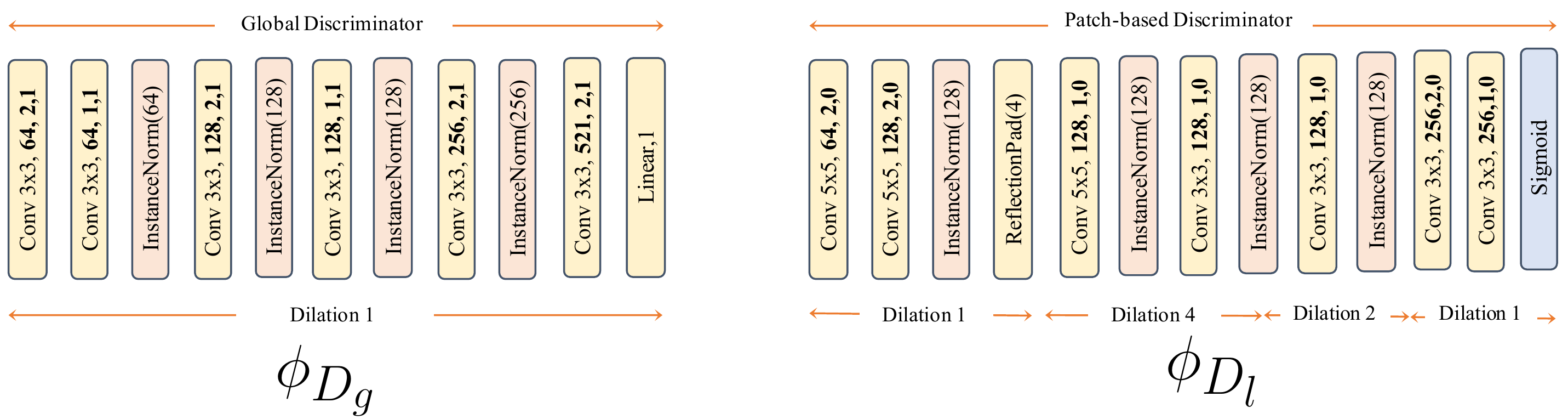}
\caption{{\bf Discriminator Architecture Details.} We show the full architecture for the two discriminators: a global, holistic $\phi_{D_{g}}$ and  a patch-based local discriminator $\phi_{D_{l}}$.}%
\label{fig:disc}
\end{figure*}

\minisection{Discriminators:} The discriminators used in training are shown in \figref{disc}. Unlike the generator, these two discriminators use always regular convolutions. After convolutions, LeakyReLU is used as activation. The patch-based discriminator is designed to yield a 50x50 receptive field in the input image.  The legend for the blocks in \figref{disc} is as follows: $\mathtt{Conv 3x3, 521, 2,1}$ indicates convolutional filters with kernel 3x3, an output with 512 channels, stride equal to 2, and a padding of 1.

